\pdfoutput=1
\documentclass[11pt]{article}
\usepackage[margin=1in]{geometry}
\usepackage{lmodern}
\usepackage[T1]{fontenc}
\usepackage{microtype}
\usepackage{graphicx}
\usepackage{booktabs}
\usepackage{array}
\usepackage{xcolor}
\usepackage{listings}
\usepackage{tikz}
\usetikzlibrary{arrows.meta,positioning,shapes.misc,fit,backgrounds}
\usepackage[numbers,sort&compress]{natbib}
\usepackage[hidelinks]{hyperref}
\usepackage{caption}
\usepackage{enumitem}
\setlist{nosep,leftmargin=*}
\newcommand{\sys}{Reels-RSI}

\title{\textbf{\sys{}:}\\[2pt]
\large Program-Level Self-Improvement for Code-Rendered Explainer Videos}

\author{%
\begin{tabular}{@{}c@{}}
\raisebox{-0.42\height}{\includegraphics[height=1.05cm]{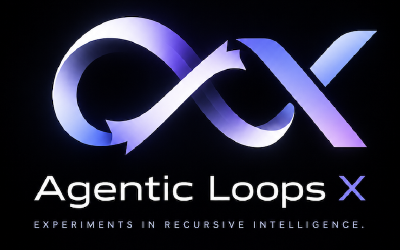}}\hspace{0.6em}%
\begin{tabular}[c]{@{}l@{}}\large\textbf{Wei Lin}\\[2pt]
\normalsize Agentic Loops X\\[1pt]
\small\url{https://github.com/agentic-loops-x}\end{tabular}
\end{tabular}
}
\date{\small October 2026 \quad\textbullet\quad Preprint \quad\textbullet\quad Code, rules, lessons and the evolve report: \url{https://github.com/agentic-loops-x/Reels-RSI}}

\begin{document}
\maketitle

\begin{abstract}
Explainer videos---a worked geometry problem, why the sky is blue, how a dynasty fell---are usually
one more target for pixel video generation. We argue they are a different kind of
artifact: a \emph{program} (script, storyboard and per-frame animation code) that a
deterministic renderer turns into frames. A program can be read, diffed, tested and audited; pixels
can only be watched. \sys{} is an open-source reel director for any agent harness; its
self-improvement happens at the program level rather than in weights or prompts. Every revision
pass is logged with a code snapshot; a retrospective pairs each check finding with the diff that
fixed it and proposes lessons; a lesson that a static check can catch is \emph{compiled into a
rule} with a must-fail and a must-pass example, so the memory is unit-tested; and a rule learned
today runs over everything made before it. A bounded, benchmark-gated step lets an agent rewrite
the director's playbook, kept only when blind pairwise judgments on held-out topics prefer it;
nothing enters memory without a human decision. Across 111 reels---a field study, two playbook
evolutions, a 2$\times$2 memory ablation, a weaker director, a 48-reel learning curve and a
judge-reliability study---we find that the rule learned in the field caught four latent defects in
frames authored before it existed and none after; that rules are an audit for a strong director
(eight unguarded reels, no violation) and a guard for a weaker one (thirty violations unguarded
versus five); that lessons shorten revision loops but do not compound or move the judge; and that
the first rewrite's 11--1 verdict was a position bias in our vote design (judges pick the
second-shown reel 62--81\% of the time), masking scripts that traded thirteen facts for craft. We
document four failure modes of the preference signal, which check caught each, and the fixes. Code, rules, lessons and data are public.
\end{abstract}

\section{Introduction}

A short explainer video has a property that sets it apart from most video: nearly everything that
matters in it is \emph{symbolic}. The numbers in a worked problem, the labels on a map, the order
of steps in a derivation, the words the narrator says---these are the content; the picture exists
to deliver them. A generated clip is judged on whether its motion looks plausible
\citep{vbench2}; an explainer is judged on whether the shaded triangle's area really is
$0.3\,\mathrm{cm}^2$ and whether the viewer saw why.

Recent systems therefore generate explainers as \emph{code}---Manim programs
\citep{theoremexplain2025,code2video2026,manimagent2026}, slide decks \citep{paper2video2025}, HTML
compositions---and reap exact text, real geometry and reproducible output. We take the consequence
one step further. If the reel is a program, then everything we know how to do with programs applies
to reels: they can be linted, diffed, tested, audited after the fact, and carried between the
models that wrote them. In particular, an agent that makes reels can \emph{improve itself at the
program level}, by accumulating executable checks over its own output, instead of only in its
prompt or its weights.

This paper presents \sys{} and the self-improvement loop built into it. The system is engineering:
a deterministic command-line pipeline (narration with per-word timings, karaoke captions, font
subsetting, procedural music and sound, map data, assembly, checks, render) that any agent
harness---Claude Code, Codex CLI, Gemini CLI, OpenCode---drives from a Markdown skill; it makes
reels in Chinese, English, Japanese, Korean, Spanish and French and dubs a finished reel into
another language. The contributions are what the program view makes possible on top of it:

\begin{enumerate}
\item \textbf{A four-tier, program-level memory} (\S\ref{sec:rsi}). Every finalize pass is logged
  with a code snapshot (L0). A retrospective pairs each check finding with the diff that fixed it
  and proposes \emph{lessons}, injected into the next reel's frame briefs and scoped to the presets,
  aspects and modes they hold for (L1). A lesson that a static check can detect is \emph{compiled
  into a rule}---a \texttt{check(doc)} function shipped with must-fail and must-pass examples,
  admitted only if it passes both (L2). A benchmark-gated step lets an agent rewrite the director's
  playbook, kept only if blind pairwise judges prefer the result on training \emph{and} held-out
  topics (L3). Every tier is written by a check or by a person; the model only proposes.
\item \textbf{Retroactive audit, and an honest account of the guard} (\S\ref{sec:audit},
  \S\ref{sec:ablation}). A rule learned from reel $n$ is a program that runs on reels $1 \ldots n-1$.
  The rule learned during our study found four latent defects in two reels delivered and reviewed
  before the system existed---defects no one had noticed---and nothing in the 21 reels made under it.
  The run logs and a rules-off ablation then show why the zero is cheap: for this director the rules
  fired three times in 58 passes, and eight reels made with no rule active contain no violation. The
  same four topics under a weaker director carry thirty violations unguarded and five guarded: the
  audit is what the program view buys a strong director, the guard is what it buys a weak one.
\item \textbf{A measured account of the preference signal} (\S\ref{sec:evolve}--\S\ref{sec:reliability}).
  One evolution round was accepted by blind pairwise judgment 11--1 on training topics and 8--4 on
  held-out topics while the absolute rubric did not move. Re-judging 170 times shows the rubric stable
  (within-reel SD 1.3) but insensitive, the pairwise verdict repeatable but leaning on a position bias:
  three judge models pick the second-shown reel 62--81\% of the time, and our own three-vote
  alternation gave the candidate the extra second slot. Judged from the disadvantaged slot the round still passes, but three of eight topics are ties and
  the remaining margin is not significant ($p=0.39$ on train). Reading all sixteen scripts shows the preferred reels dropping
  thirteen baseline facts over six topics while adding six. We report four failure modes of the
  signal---a judge misreading that became a proposed lesson, a proposer writing for the sampling
  schedule, craft bought with facts, and our own vote design---and the check that caught each.
\end{enumerate}

We are explicit about scale. This is a field study of one system during its own release testing:
23 reels in the main study, 16 in the ablation, 8 under a weaker director, 48 in a learning curve, 16 in a second evolution round; one model family as director, judge and retrospective
model, one evolution round, no human preference study. It establishes what the mechanisms catch and
what they miss, not effect sizes. A companion benchmark for explainer-reel systems, with outcome tests
on the final video, is in preparation and is deliberately kept out of this paper.

\section{Explainer reels as programs}
\label{sec:modality}

\paragraph{Definition.} In \sys{} a reel is a directory. \texttt{SCRIPT.md} holds the spoken lines,
one block per frame; \texttt{STORYBOARD.md} holds per-frame scene, duration, transition, route and
status; \path{compositions/frames/} holds one \texttt{.html} composition per frame---HTML, SVG,
Canvas or Three.js driven by a single paused GSAP timeline---plus optional shared kits
(\texttt{assets/*.js}) and overlays; the voice, caption and music assets are derived from these by
the pipeline. Rendering is deterministic: a headless browser seeks each composition's timeline to
successive frame times and FFmpeg encodes \citep{hyperframes}. Re-running \texttt{finalize} on a copy
of a finished reel reproduces its assembled composition byte for byte.

\paragraph{What the program view buys.} Table~\ref{tab:modality} contrasts the two routes to an
explainer video. The right-hand column makes no claim about which route produces better pictures;
it lists what becomes \emph{possible} once the artifact is source code, and \S\ref{sec:rsi} uses
each line.

\begin{table}[t]
\centering\small
\begin{tabular}{p{2.9cm}p{5.1cm}p{6.3cm}}
\toprule
& Pixel generation & Program rendering (\sys{}, Manim-based agents) \\
\midrule
Text and numbers & sampled; may be wrong or illegible & exact strings in the program \\
Correctness check & watch the video (human or VLM) & read the program; the render is deterministic \\
Targeted edit & regenerate & change one value, re-render one frame \\
Learning a mistake & fine-tune or re-prompt & write a check over the program text \\
Applying it to old output & impossible & run the check over the corpus \\
Portability across models & none & the check is code; the model is irrelevant \\
Reproducibility & seed-dependent & byte-identical \\
\bottomrule
\end{tabular}
\caption{Two routes to an explainer video.}
\label{tab:modality}
\end{table}

\paragraph{Three layers of verification.} The artifact exists at three levels---program text,
assembled composition, rendered pixels---and can be checked at each, with rising cost and falling
certainty. \emph{Static} rules run over the program text (\S\ref{sec:rules}): exact for the pattern
they encode, free to run. \emph{Dynamic} checks run on the assembled composition---the renderer's own
lint, runtime, layout and contrast checks---exact for what they cover. \emph{Judged} assessment lets
a vision-language model score or compare sampled frames: cheap and informative, and, as
\S\ref{sec:evolve} shows, not to be trusted alone. The design principle of the loop follows from
this ordering: a learned judge may \emph{rank}; admission to memory requires a deterministic
check or a person.

\section{\sys{}}
\label{sec:system}

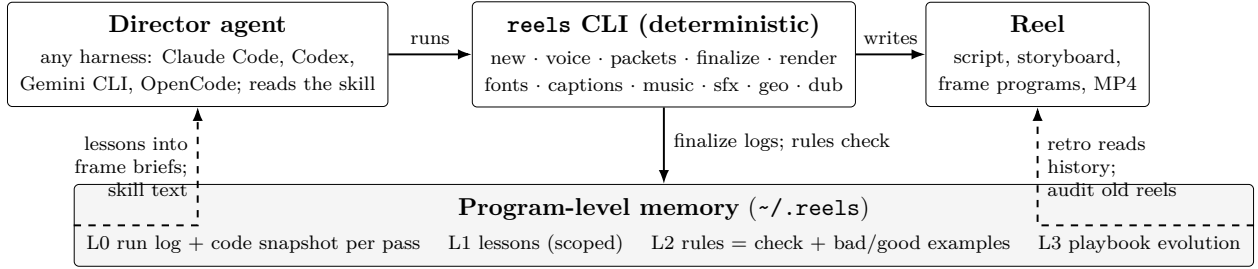
\begin{figure}[t]
\centering
\resizebox{\textwidth}{!}{\begin{tikzpicture}[
  box/.style={draw,rounded corners=2pt,align=center,inner sep=5pt,font=\small},
  mem/.style={box,fill=black!4},
  arr/.style={-{Latex[length=2mm]},thick},
  feed/.style={-{Latex[length=2mm]},thick,dashed}]
\node[box,minimum width=4.2cm,minimum height=1.5cm] (agent) {\textbf{Director agent}\\[1pt]\scriptsize any harness: Claude Code, Codex,\\[-1pt]\scriptsize Gemini CLI, OpenCode; reads the skill};
\node[box,right=1.2cm of agent,minimum width=5.4cm,minimum height=1.5cm] (cli) {\textbf{\texttt{reels} CLI (deterministic)}\\[1pt]\scriptsize new $\cdot$ voice $\cdot$ packets $\cdot$ finalize $\cdot$ render\\[-1pt]\scriptsize fonts $\cdot$ captions $\cdot$ music $\cdot$ sfx $\cdot$ geo $\cdot$ dub};
\node[box,right=1.0cm of cli,minimum height=1.5cm] (reel) {\textbf{Reel}\\[1pt]\scriptsize script, storyboard,\\[-1pt]\scriptsize frame programs, MP4};
\node[mem,below=1.1cm of cli,minimum width=11.6cm] (mem) {\textbf{Program-level memory} (\texttt{\textasciitilde/.reels})\\[2pt]
\scriptsize L0 run log + code snapshot per pass \quad L1 lessons (scoped) \quad L2 rules = check + bad/good examples \quad L3 playbook evolution};
\draw[arr] (agent) -- node[above,font=\scriptsize]{runs} (cli);
\draw[arr] (cli) -- node[above,font=\scriptsize]{writes} (reel);
\draw[arr] (cli.south) -- node[right,font=\scriptsize,pos=.45]{finalize logs; rules check} (cli.south |- mem.north);
\draw[feed] (mem.west) -| node[left,font=\scriptsize,pos=.75,align=right]{lessons into\\frame briefs;\\skill text} (agent.south);
\draw[feed] (mem.east) -| node[right,font=\scriptsize,pos=.75,align=left]{retro reads\\history;\\audit old reels} (reel.south);
\end{tikzpicture}}
\caption{\sys{}. The creative half (script, storyboard, frame code, review) belongs to the agent;
the deterministic half is a CLI the agent calls. Memory sits beside both and is written only by
checks and by human-gated acceptance.}
\label{fig:arch}
\end{figure}

\sys{} (Fig.~\ref{fig:arch}) separates the creative work from everything that must be
reproducible. The \textbf{skill} is about 1{,}000 lines of Markdown: a ten-step director's handbook
with references on script and storyboard, frame construction, a quality bar, worked problems,
history, visual styles and techniques. Any agent that can read files and run commands can follow
it. The \textbf{CLI} (about 5{,}000 lines of Python on the standard library, a TTS client and font
tools) owns narration with per-word timings, caption building with CJK-aware grouping and bilingual
lines, font subsetting so the headless renderer never shows missing glyphs, procedural music beds
and sound effects cut to the real narration, Natural Earth map layers, Wikimedia Commons images with
credits, stroke-order data for Chinese characters, assembly, overlays, transitions, the renderer's
checks, snapshots and render. Five visual presets (deep-space, notebook, ink, atlas, chalk) each
ship a design spec and a caption skin; chalk ships a drawing kit.

\paragraph{Models are interchangeable at two seams.} The director runs inside whichever agent CLI
the user already has; \texttt{reels make "\ldots"} launches a headless session. The judge and
retrospective roles are one-shot calls, named \texttt{provider:model}, over an OpenAI-compatible or
Anthropic HTTP API or through a logged-in agent CLI; twelve providers and custom endpoints are
configured in a TOML file. Nothing in memory depends on the model that produced it.

\paragraph{Languages.} A reel's language is detected from its script (Hangul, kana, Han, else Latin
scored by stop words); the voice, caption cutting (grouped by character for Chinese and Japanese, by
word otherwise), glyph source (Noto SC, JP or KR, fetched on first use) and subtitle export follow.
\texttt{reels dub} translates a finished reel's spoken lines and every on-screen string in one model
call, re-voices each line fitted to the original line's duration so the frames stay in step, and
re-renders; the geometry lesson of \S\ref{sec:study} exists in six languages this way.

\section{Program-level self-improvement}
\label{sec:rsi}

The loop has four tiers. The invariant across them: a model may propose, and only a check or a
person may commit.

\subsection{L0: every pass is evidence}
\texttt{reels finalize} assembles the reel and runs every check---the renderer's lint, runtime,
layout and contrast checks, then every \sys{} rule. Each pass appends a record to
\texttt{.reels/runs.jsonl} (findings with codes, lines and messages; frame count; duration) and
snapshots the frame sources to \texttt{.reels/history/NNN/}. Renders, scores and viewer feedback
(\texttt{reels feedback "the labels are too small" --frame 04}) append to the same log. Apart from the
judge's score, no model writes this tier; it is the raw material the next tiers read.

\subsection{L1: lessons, paired with the diff that earned them}
\texttt{reels retro} compiles an \emph{evidence file} from the log: per pass, which findings were new
and which were fixed; for every fixed finding, the unified diff between the snapshot in which it was
present and the one in which it was gone, labelled with what it fixed; the viewer's comments; the
judge's notes; and everything already known---accepted lessons and existing rules---so nothing is
proposed twice. A retrospective model reads the evidence and proposes at most five lessons, each with
a kind (\texttt{doc}, \texttt{rule}, \texttt{kit}, \texttt{taste}), a scope (director, frame, solve,
history, \ldots) and an optional \emph{when} condition such as \texttt{preset=chalk},
\texttt{aspect=9:16} or \texttt{mode=solve}. Lessons land in an inbox, which is inert. On
\texttt{accept} they are injected into the briefs of every future frame worker---but only the lessons
whose conditions the new reel satisfies, so a 16:9 science reel is not told how to draw chalk boxes
on a portrait board.

\subsection{L2: a lesson that can be checked becomes a rule}
\label{sec:rules}
A rule is a directory: \texttt{rule.py} exposing \texttt{RULE} (id, severity, lesson, fix) and
\texttt{check(doc)} over a frame's text, plus \texttt{bad.*} files that must produce a finding and
\texttt{good.*} files that must not. \texttt{reels rules test} runs every rule against its examples;
\texttt{reels lessons accept --rule-dir} refuses a rule lesson whose rule fails them. Rules run on
every \texttt{finalize} and \texttt{lint}; once admitted, the mistake cannot recur, whichever model is
directing. Listing~\ref{lst:rule} shows the rule learned during the study; Appendix~\ref{app:rules}
lists the rest.

\begin{lstlisting}[language=Python,caption={The rule compiled from a lesson during the study. Its must-fail example is the
original chalk-dust helper from the reel in which the defect appeared; the must-pass example adds
\texttt{immediateRender:\ false} and a fade-in from zero, which is always safe.},label={lst:rule}]
RULE = {"id": "visible-from-state", "severity": "warning",
        "kinds": ["frame", "overlay", "kit"],
        "lesson": "fromTo renders its from-state at build time: a tween that starts "
                  "visible (opacity > 0) and fades to 0 later shows the element before its cue",
        "fix": "Add immediateRender: false to the to-vars"}

def check(doc):
    out = []
    for idx, args in H.calls(doc.text, "fromTo"):
        if len(args) < 4: continue                     # no position: starts at t=0 anyway
        frm, to = args[1], args[2]
        a, b = num(frm, "opacity"), num(to, "opacity")
        if a is not None and a > 0 and b == 0 and "immediateRender" not in to:
            out.append(H.Finding(RULE["id"], RULE["severity"], doc.rel,
                                 H.line_of(doc.text, idx),
                                 "fromTo starts visible and fades out - visible before its cue"))
    return out
\end{lstlisting}

The line between L1 and L2 is the line between advice and enforcement. A doc lesson says ``labels
inside coloured regions go at the region's widest part''; a model may or may not follow it, and no
one can tell afterwards whether it did. A rule says ``a \texttt{fromTo} that starts visible and fades
out without \texttt{immediateRender:false} is a finding''; no model can skip it, and its own examples
say whether it is right. Only mistakes detectable from program text can become rules---a real limit
(\S\ref{sec:limits}) and the reason L1 exists.

\subsection{L3: the playbook rewrites itself, under a gate it cannot move}
\label{sec:evolve-method}
\texttt{reels evolve} runs one round. (1)~The current skill makes the four training topics of a
fixed eight-topic benchmark (science, history, a worked problem, an English explainer). (2)~An agent
reads the baseline's judge notes, check findings and the lesson inbox and makes at most three edits
to a \emph{copy} of the skill's Markdown, explaining each in a changelog. (3)~The copy makes the same
topics. (4)~Per topic, a vision judge sees both reels blind, three votes with the order swapped
between votes; the candidate must win most votes, render at least as many reels, and lose no topic by
more than ten composite points. (5)~Both versions make the four held-out topics, and the candidate
must win at least half the votes there. (6)~A report is written with per-topic votes, the judge's
reasons and a side-by-side image of both reels; nothing changes until a person runs
\texttt{evolve apply}. The proposer may edit only the skill text. The pipeline, the rules, the
presets, the judge's rubric and the benchmark topics are out of its reach, and it is told to record
tool defects it discovers in a separate file for a developer rather than work around them in prose.

\section{Field study}
\label{sec:study}

\subsection{Setting}
The data are the reels \sys{} made during its own release testing in October 2026. Claude Code was
the director (Opus for the two reels made with a human reviewer, Sonnet for the headless benchmark
reels); \texttt{claude-cli:sonnet} was the judge and the retrospective model. Table~\ref{tab:corpus}
lists the corpus: two reels predate the system (made with its predecessor skill, before any rule
existed); two were made from scratch with a person reviewing at three gates (plan, look-dev,
final); seventeen were made headlessly from one sentence each in the benchmark and evolve runs; two
are dubs. A 30-second headless reel cost \$0.56--1.29 and 230--584~s of wall time (15--34 agent
turns); judging a reel takes about 30~s.

\begin{table}[t]
\centering\small
\begin{tabular}{lp{4.3cm}rrlp{3.4cm}}
\toprule
Group & Reels & Frames & Length & Lang.\ / aspect & Made by \\
\midrule
Pre-existing & 2 (Yuan dynasty; why the sky is blue) & 19 & 66--96 s & zh, 16:9 & predecessor skill, no rules \\
Reviewed & 2 (meeting problem; textbook geometry from a photo) & 16 & 55--84 s & zh, 9:16 & Opus, reviewer at 3 gates \\
Headless & 17 (8 topics $\times$ 2 skills, +1) & 83 & 26--33 s & zh/en, 16:9 & Sonnet, one sentence in \\
Dubbed & 2 (the geometry reel in en, ja) & 18 & 84 s & en, ja, 9:16 & \texttt{reels dub} \\
\midrule
Total & 23 & 136 & & zh, en, ja & \\
\bottomrule
\end{tabular}
\caption{The corpus. Dubs of the geometry reel into Spanish, French and Korean exist but were not
rendered at the time of writing and are excluded.}
\label{tab:corpus}
\end{table}

\paragraph{Compute.} Everything in this paper was made on one laptop and one consumer subscription.
The 105 headless reels across the benchmark, two evolution rounds, ablation, weaker-director and
learning-curve runs cost \$82.72 in metered agent time (\$0.35--1.45 per 30-s reel); the two reviewed reels were made
in an interactive session and the roughly 400 judge calls ran on the subscription, unmetered. No
experiment needed more than one night.

\subsection{Within a reel, the review loop converges}
Every reel reached zero check errors before render. The two reviewed reels needed 7 and 5 finalize
passes; their warning counts went $5 \rightarrow 12 \rightarrow 2 \rightarrow 2 \rightarrow 1
\rightarrow 1 \rightarrow 1$ and $1 \rightarrow 1 \rightarrow 3 \rightarrow 2 \rightarrow 1$ (the rise
in pass~2 is new frames being added). The headless reels needed 1--6 passes (median 2); the hardest,
a history reel, went $3, 4, 3, 0$ errors over four passes. The two reviewed reels also exposed
thirteen defects in the tool and the skill---a caption skin with 1.27:1 contrast, placeholders
missing for unbuilt frames so voice and captions drifted, a map-winding rule that flagged a car
outline as a polygon---each fixed in the tool rather than patched in the reel, which is the point of recording where a fix went.

\subsection{Lessons}
\label{sec:lessons}
Retrospectives on the two reviewed reels, and the agents' own retrospectives after each headless
reel, produced 56 proposed lessons. A person accepted 15, rejected 1 and left 40 in the inbox. Of the
15, thirteen are \texttt{doc} lessons (e.g.\ ``budget 3.9 Chinese characters per second for the
teaching voice, not 4.5--5''; ``a mark that underlines a word rides on the text element, never on
guessed coordinates''), one is a \texttt{kit} lesson (every reel gets a drawing kit; the chalk kit now
ships with the preset) and one is a \texttt{rule} lesson (\S\ref{sec:audit}). Two carry \emph{when}
conditions (\texttt{preset=chalk}, \texttt{aspect=9:16}); before conditions existed, a 16:9
deep-space reel had been handed both. The full list is in Appendix~\ref{app:lessons}.

The rejected lesson matters more than the accepted ones. The retrospective model had turned a judge's
note---``frame 09 is rendered at the wrong aspect, 1080$\times$640''---into ``verify every frame's
render dimensions before finalizing''. The frame was 1080$\times$1920; the judge had been shown a
sample sheet downscaled until each portrait frame was about 200 pixels wide. Had lessons been
accepted automatically, a false constraint would have reached every subsequent reel. The
retrospective prompt now ranks evidence (check findings and viewer words above judge notes), and sample
sheets are sized per aspect.

\subsection{Lesson to rule, and the audit it enables}
\label{sec:audit}
In the first reviewed reel, a chalk-dust puff showed stray specks before its cue. The cause is a GSAP
semantic: \texttt{fromTo} renders its from-state at build time, so a tween from opacity 0.9 to 0,
placed later in the timeline, shows the element from the first frame. The fix is one property. The
lesson became the rule of Listing~\ref{lst:rule}, with the original helper as its must-fail example.
We then ran all thirteen rules over the whole corpus (Table~\ref{tab:audit}).

\begin{table}[t]
\centering\small
\begin{tabular}{lrrrl}
\toprule
Reels & Frames & Authored & Findings & Where \\
\midrule
2 pre-existing & 19 & before the rule & 4 & Yuan: frames 06, 09, 11; sky: frame 04 \\
21 later & 117 & with the rule active & 0 & \\
\bottomrule
\end{tabular}
\caption{Retroactive audit with \texttt{visible-from-state}. The other twelve rules were ported from
the predecessor's checker, under which the pre-existing reels were made, and produced no findings on
any reel; the signal here is specific to the rule learned later.}
\label{tab:audit}
\end{table}

All four findings are defects by construction: each is a fade-out tween whose visible from-state
renders at $t=0$. The sky reel's finding was confirmed in a rendered frame---a ripple ring that should
appear at its cue is already standing in the picture at 21.0~s. Both pre-existing reels had been
delivered and reviewed by a person, and nobody had noticed. This is the line of Table~\ref{tab:modality}
that matters most: a lesson learned on reel $n$, once compiled, is a program, and programs run on reels
$1 \ldots n-1$. A prose lesson could not have found these, and a judge was never asked.

The guard side of the same table is the zero, and it needs a caveat the run logs supply. Across the
58 finalize passes of the 19 reels with logs, the thirteen rules fired only three times
(\texttt{repeat-fromto} twice, \texttt{dasharray-attr} once; each fixed before the next pass), while
the renderer's own checks fired 184 times (30 errors, 154 warnings). The agents that wrote the 117
frames rarely made the mistakes the rules encode. So the zero is mostly the model's doing, and the
rules are a backstop that was needed three times---the ablation in \S\ref{sec:ablation} measures what
appears when the backstop is removed. Over every reel with a log---92 reels, 260 finalize passes---the thirteen rules fired 25 times, 18 of
them in one weak-director reel (\S\ref{sec:ablation}); nine rules, \texttt{visible-from-state} among
them, never fired while a reel was being authored. The retroactive audit is the stronger claim: it
does not depend on how often a mistake is made, only on the corpus being code.

\subsection{One evolution round}
\label{sec:evolve}
The round ran overnight on a consumer subscription, paused once for two hours when the quota ran
out, and resumed from its state file: sixteen headless reels and about forty judge calls, \$12.75 in
agent cost. The proposer made three edits (a 59-line patch): checkable minimums for hero size and
camera motion; ``mechanism recipes'' (show travel and effect; do not label them); and board legibility
for worked problems. It also traced a broken 16:9 board in a chalk reel to a hard-coded portrait
canvas in the shipped kit---a code defect, reported in its tool-bug file and fixed by hand.

\begin{table}[t]
\centering\small
\begin{tabular}{lcc|cc}
\toprule
 & \multicolumn{2}{c|}{Absolute composite (0--100)} & \multicolumn{2}{c}{Blind pairwise votes} \\
 & baseline & candidate & baseline & candidate \\
\midrule
Train (4 topics $\times$ 3 votes) & 75.3 & 74.9 & 1 & \textbf{11} \\
Holdout (4 topics $\times$ 3 votes) & 72.8 & 73.5 & 4 & \textbf{8} \\
\bottomrule
\end{tabular}
\caption{The first evolve round. Composite $= 0.6 \cdot$ judge rubric $+\, 0.4 \cdot$ deterministic
score; the deterministic part was at or near 100 for both versions, so the absolute columns are an
affine rescaling of the judge's 1--5 rubric. Every held-out topic split 2--1.}
\label{tab:evolve}
\end{table}

\begin{figure}[t]
\centering
\includegraphics[width=\textwidth]{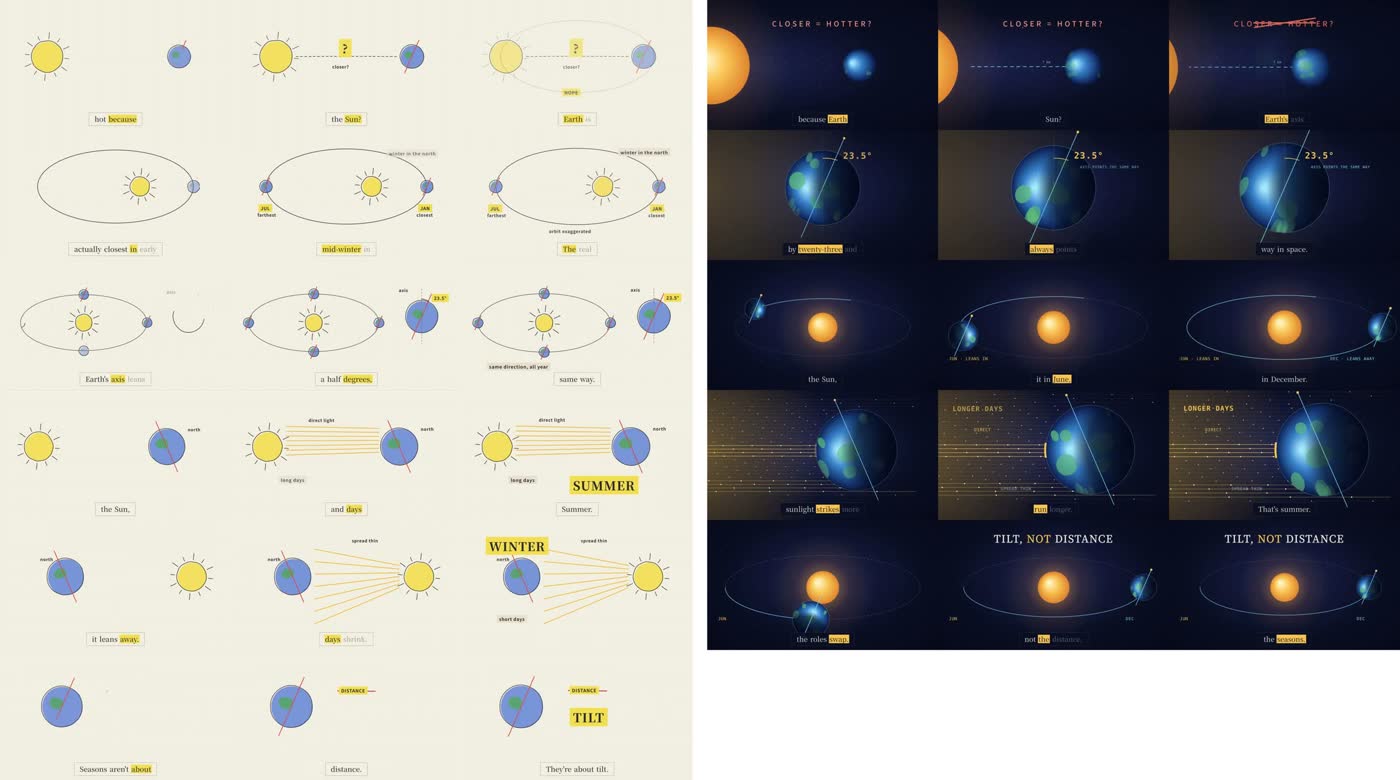}
\caption{\emph{Why do we have seasons?} under the baseline skill (left) and the evolved skill
(right): same topic, same model; each row is one frame sampled at 30, 60 and 92\% of its duration.
The judge preferred the right 3--0.}
\label{fig:seasons}
\end{figure}

Table~\ref{tab:evolve} is why the verdict is pairwise. The same judge, asked for absolute rubric
scores, moved by half a point; asked which of two reels on the same topic is better, it preferred the
candidate on 11 of 12 training votes and 8 of 12 held-out votes. Fig.~\ref{fig:seasons} shows the
largest change. The round was accepted by the gate and, after a person inspected the side-by-side
images, applied. \S\ref{sec:reliability} re-examines both numbers.

\subsection{Four ways the preference signal misled the loop}
\label{sec:hacking}
We record these because each happened, each was caught by a different check, and a loop without that
check would have absorbed it (Table~\ref{tab:hacking}). The fourth was found by checking the harness's
own verdict, which is the same discipline turned on ourselves.

\begin{table}[t]
\centering\small
\begin{tabular}{p{3.6cm}p{5.7cm}p{5.3cm}}
\toprule
Failure & What the signal produced & Gate that caught it \\
\midrule
(a) Judge misreading & a proposed lesson asserting a non-existent render bug & human \texttt{accept}; then sample sheets sized per aspect, evidence ranked \\
(b) Writing for the sampler & a playbook edit about ``the 30/60/92\,\% samples'' & human \texttt{apply}; then the proposer prompt forbids sampling-directed text \\
(c) Craft bought with facts & preferred reels whose narration drops 13 baseline facts over 6 topics & none in the loop; found by reading the scripts \\
(d) Position bias in our own vote design & a 3-vote alternation that gave the candidate the second slot twice; judges pick the second-shown reel 62--81\,\% of the time ($p<10^{-5}$ for the round's judge) & the reliability study (\S\ref{sec:reliability}); the gate now judges each side from its disadvantaged slot \\
\bottomrule
\end{tabular}
\caption{Four failure modes of the judge signal observed around one round, and where each stopped.}
\label{tab:hacking}
\end{table}

\paragraph{(a) A judge misreading became a proposed lesson.} Described in \S\ref{sec:lessons}: the
retrospective model treated a judge note as fact. Caught at the human \texttt{accept} gate; prevented
upstream afterwards.

\paragraph{(b) The proposer wrote for the sampler.} One edit in the accepted patch read ``the
30/60/92\,\% samples must differ by camera position''---an instruction about the judge's sampling
schedule, not about what a viewer sees. Caught at the human \texttt{apply} gate and rewritten as
``keep the picture changing until the narration ends''; the proposer prompt now forbids instructions
aimed at sampling. This is the ``teaching to the test'' of \citet{notanoracle2026} in a creative
setting, produced within a single round.

\begin{figure}[t]
\centering
\includegraphics[width=\textwidth]{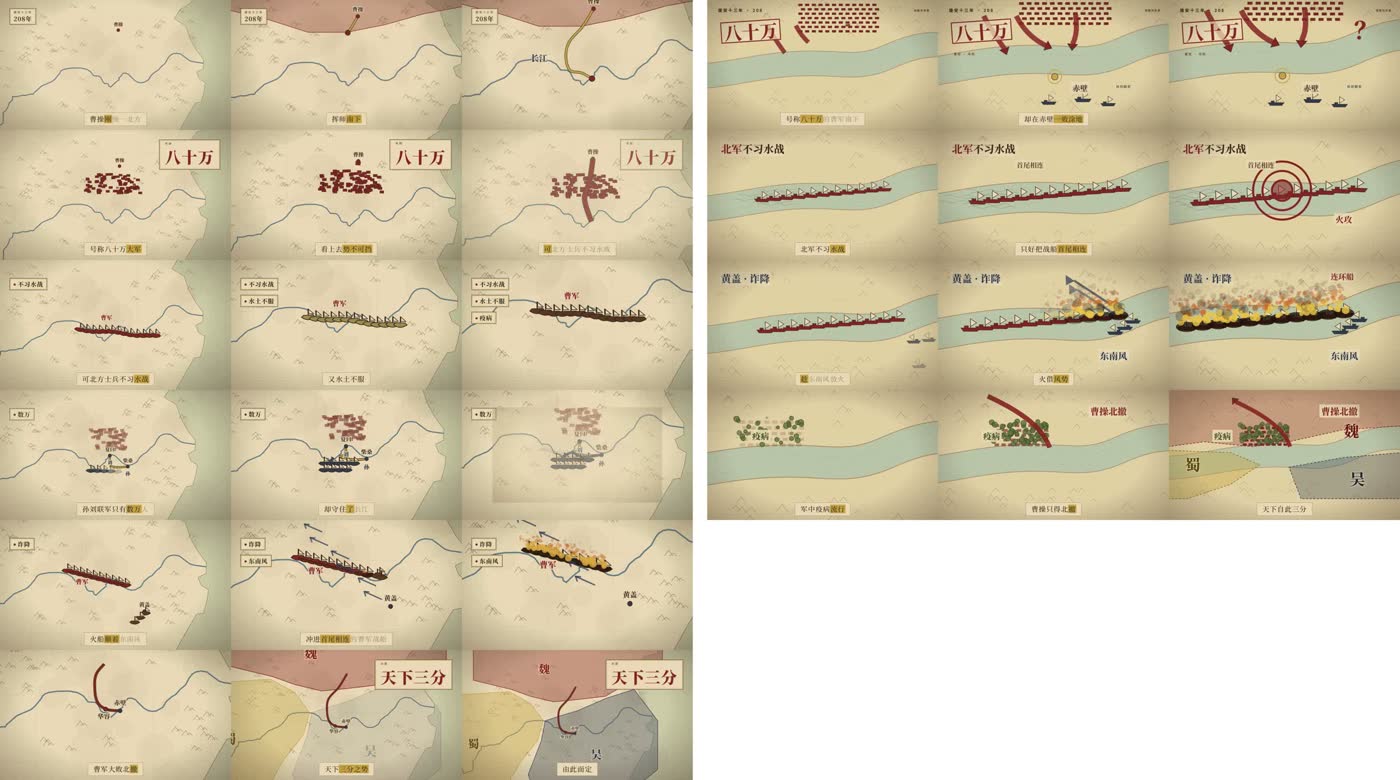}
\caption{Red Cliffs, baseline (left) and evolved (right). The baseline's fourth row shows the
Sun--Liu alliance holding the river; the evolved reel has no such frame, and its narration never
names the alliance. The judge preferred the right 2--1 for its larger, animated fire scene.}
\label{fig:chibi}
\end{figure}

\paragraph{(c) Craft was bought with facts.} On the Red Cliffs topic the candidate reel is visibly
richer (Fig.~\ref{fig:chibi}) and won 2--1. Its script lists three reasons Cao Cao lost and never
names the Sun--Liu alliance that beat him; the baseline devotes a frame to it. We then read all
sixteen scripts (Table~\ref{tab:facts}). The candidate drops at least one fact the baseline states on
six of eight topics---the January perihelion that the seasons reel exists to correct, the 12\,h\,25\,min
tide interval, Zhang Qian and Emperor Wu on the Silk Road, the 40$^\circ$ violet angle, the buoyancy
numbers---and adds a new one on five. Script length is unchanged ($1{,}092 \rightarrow 1{,}063$
characters, $-2.7\%$): the evolved skill, whose three edits are all about the picture, substitutes
facts rather than compressing them, and the judge, which sees stills and a storyboard digest, cannot
tell. Nothing in the evolve verdict looks at facts. Because the reel is a program, each omission is a
one-line search over \texttt{SCRIPT.md}, which is how we found them---but the loop did not run that
search, because no rule asked for it. A rule cannot know which facts a topic requires; an outcome test
per topic can, which is what the companion benchmark adds and why we keep it separate.

\paragraph{(d) Our own vote design leaned on the judge's position bias.} \S\ref{sec:reliability}: the
judge picks the second-shown reel four times in five, and the three-vote alternation handed the
candidate that slot twice per topic. Judged only from the disadvantaged slot, three of the eight topics it reported as wins are ties and
the remaining preference is not significant at this size. Caught by re-running the judging; the gate
now uses the position-robust rule.

\begin{table}[t]
\centering\small
\begin{tabular}{lp{6.2cm}p{4.6cm}c}
\toprule
Topic & Stated by the baseline, absent from the candidate & Added by the candidate & Votes \\
\midrule
sci-rainbow & white light split into seven colours; violet at $\sim$40$^\circ$; the viewer at the apex of a cone & --- & 0--3 \\
hist-chibi & Cao Cao had just unified the north; the Sun--Liu alliance held the river; soldiers unacclimatised & chained ships became the fire attack's target & 1--2 \\
solve-chicken-rabbit & --- & the general rule (assume all chickens; halve the missing feet) & 0--3 \\
en-seasons & Earth is closest to the Sun in early January; in winter light spreads thin, days shrink & --- & 0--3 \\
sci-tides & the two high tides are $\sim$12\,h\,25\,min apart because the Moon moves on & the Moon's pull differs across the Earth & 1--2 \\
hist-silk-road & Emperor Wu of Han; Zhang Qian as envoy; Buddhism travelling east & Yumen Pass; porcelain west, glass east & 1--2 \\
solve-buoyancy & the worked numbers (1.3\,N vs 10\,N; 1\,000\,cm$^3$) & Archimedes' principle stated & 1--2 \\
poem-jingyesi & (scene-setting only) & --- & 1--2 \\
\bottomrule
\end{tabular}
\caption{Fact coverage of the two scripts per topic, read by hand (a fact = a number, name, date or
causal step). Votes are baseline--candidate from the round. The candidate drops thirteen baseline facts
over six topics and adds six over five; the judge preferred it on all eight.}
\label{tab:facts}
\end{table}

\subsection{Memory ablation}
\label{sec:ablation}
Does the memory change the reels? We made the four training topics four more times with the
current (evolved) skill and the same director, crossing rules \{on, off\} with lessons \{on, off\}:
\texttt{REELS\_RULES=off} removes every rule from \texttt{finalize} and \texttt{lint}; an empty
\texttt{REELS\_HOME} removes the fifteen accepted lessons from the frame briefs. All sixteen reels
rendered. Afterwards we ran all thirteen rules offline over every reel, whatever switch it was made
with, and had the judge compare conditions pairwise, position-balanced (four votes per topic, each
order twice). Table~\ref{tab:ablation} has the numbers.

\begin{table}[t]
\centering\small
\begin{tabular}{lcc|cccc|cc}
\toprule
 & \multicolumn{2}{c|}{memory} & rule violations & finalize & renderer warn. & judge & \multicolumn{2}{c}{pairwise, 4 topics $\times$ 4 votes} \\
Condition & rules & lessons & in the output & passes & at last pass & rubric & vs lessons off & vs rules off \\
\midrule
A & on & on & 0 & 17 & 0 & 58.7 & 8--8 (vs C) & 5--11 (vs B) \\
B & off & on & 0$^\dagger$ & 10 & 0 & 60.5 & 7--9 (vs D) & --- \\
C & on & off & 0 & 11 & 3 & 59.8 & --- & 5--11 (vs D) \\
D & off & off & 0$^\dagger$ & 12 & 3 & 58.0 & --- & --- \\
\bottomrule
\end{tabular}
\caption{The 2$\times$2 memory ablation, four training topics each, Sonnet directing. $^\dagger$The
offline audit of the eight unguarded reels raised one finding, a counter-clockwise ring in a Red Cliffs
map; the region renders correctly because that kit projects points into a plain SVG path, where ring
direction is irrelevant, so it was a false positive of \texttt{polygon-winding}, which now fires only
where a ring becomes GeoJSON and ships this kit's pattern as a must-pass example.}
\label{tab:ablation}
\end{table}

\paragraph{The guard was not needed.} Eight reels made with no rule active contain no true violation
of any of the thirteen rules. For this director, the mistakes the rules encode are ones it no longer
makes---consistent with the three firings in 58 passes of \S\ref{sec:audit}. The one finding the audit
did raise was a false positive, which is the other thing an audit over a corpus of programs finds:
imprecise rules. It cost a two-line change to the rule's gate and a new must-pass example, after which
every rule still passes its examples.

\paragraph{Lessons left no trace the judge could see, and a small one the renderer could.} With rules
on, the reels made with lessons and without split the pairwise votes 8--8; with rules off, 7--9
(pooled 15--17, $p=0.86$). The
rubric means are within two points. The only consistent difference is at the last finalize pass: both
lesson-less conditions ended with three renderer warnings outstanding, both conditions with lessons
with none, which is the kind of thing a brief about caption bands and label placement would affect.
We read this as: fifteen lessons, most of them about chalk-board geometry reels, do not move a judge
watching four science and history stills; a lesson earns its place on the reels it was learned from.

\paragraph{Rules on were judged slightly worse.} Both rules-on conditions lost their pairwise
comparison to the rules-off twin 5--11 (pooled 10--22, $p=0.05$). With four topics we do not read a mechanism into it---no
rule fired in any of these sixteen reels, so the switch changed nothing the agent saw except the
absence of a check step---and with the second-shown reel still picked 58\% of the time even under the
balanced design, 5--11 over four topics is within what position alone produces. It is, at least, not
evidence that the guard improves craft.

\paragraph{A weaker director needs the guard.} The sixteen Sonnet reels say the rules were not needed
by that director; they do not say the rules are not needed. We repeated the four training topics with
\texttt{claude-cli:haiku} directing, rules on and rules off, lessons on (Table~\ref{tab:haiku}). The
weaker director makes exactly the mistakes the rules encode: with the rules off, its four reels carry
30 violations---nine frames with Chinese text and no shipped font, fifteen unregistered timelines, six
nondeterministic animations---and nothing flagged them. With the rules on, the same director tripped
them 18 times during authoring and fixed most: five violations remained, all in one reel that it
rendered with fifteen renderer errors still outstanding. (On this machine the missing font-faces do
not show, because the authors' Chrome has system CJK fonts; the reference Linux renderer has none and
would draw boxes.) The reels themselves are poor by every signal---deterministic score 45--58, several
not scorable by the judge---which is the point: a director that is not yet good is the one the guard is
for, and the audit found what the guard was switched off from catching. Thirty violations in four
reels against zero in eight is the difference between the two directors, not between two settings.

\begin{table}[t]
\centering\small
\begin{tabular}{lcrrrrrc}
\toprule
Director & rules & firings & violations left & passes & renderer errors left & det.\ score & cost \\
\midrule
haiku & on & 18 & 5 (1 reel) & 7 & 48 & 45 & \$2.58 \\
haiku & off & 0 & 30 (2 reels) & 4 & 67 & 58 & \$2.00 \\
sonnet (Table~\ref{tab:ablation}) & on & 0 & 0 & 28 & 0 & 100 & \$6.32 \\
sonnet (Table~\ref{tab:ablation}) & off & --- & 0 & 22 & 0 & 100 & \$7.29 \\
\bottomrule
\end{tabular}
\caption{The same four topics under a weaker director. ``Firings'' are rule findings raised at finalize
while authoring; ``violations left'' are what all thirteen rules find in the delivered reels. Sonnet
rows pool the two lessons conditions.}
\label{tab:haiku}
\end{table}

\subsection{Does the memory grow? Forty-eight more reels}
\label{sec:curve}
The ablation compared four reels per cell with a fixed store of fifteen lessons. To ask whether the
L1 tier \emph{accumulates}, we wrote 24 new topics (six each of science, history, worked problem and
English explainer, in that repeating order), and made each twice with the same director (Sonnet, 30-s
draft, rules on): once in an arm whose lesson store starts empty and \emph{auto-accepts} every
\texttt{doc}, \texttt{kit} and \texttt{taste} lesson the agent files after each reel, so reel $k$ sees
what reels $1 \ldots k-1$ taught; once in an arm where the same lessons are filed but never accepted.
The auto-accept arm is an explicit departure from the human gate, labelled as such, because a curve
needs a loop that closes without a person. All 48 reels rendered, in 4.3 hours for \$37.25; no rule
fired in the lessons arm, four times in the other (each fixed); the offline audit finds no violation in
any of the 48. The store grew from 3 lessons after the first reel to 51 after the last.

\begin{figure}[t]
\centering
\includegraphics[width=\textwidth]{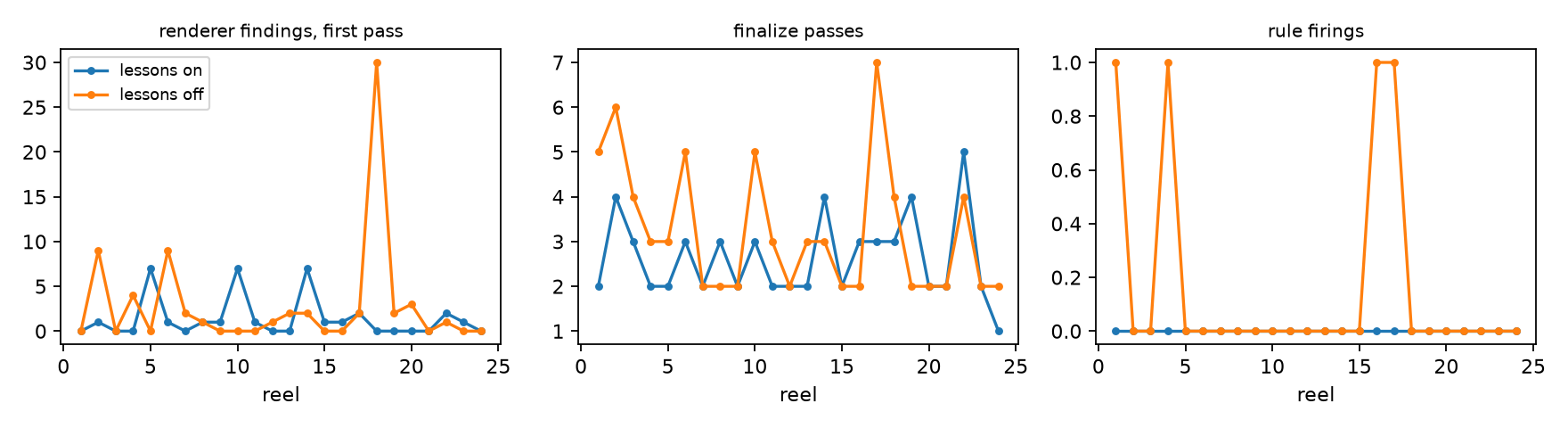}
\caption{Per reel, in order: renderer findings at the first finalize pass, finalize passes to a clean
build, and rule firings; lessons accumulating (blue) against none (orange). Topics cycle science,
history, problem, English, which is the period in the orange spikes.}
\label{fig:curve}
\end{figure}

\paragraph{Lessons lower the review cost; they do not keep lowering it.} Paired by topic, the arm
with lessons needed fewer finalize passes (2.62 vs 3.21; fewer on 12 topics, more on 5, equal on 7;
sign-flip permutation $p=0.069$) and had fewer renderer findings at its first pass (errors 1.04 vs
2.21, warnings 1.38 vs 2.83; both heavy-tailed and not significant at $n=24$, $p=0.29$ and $0.41$).
The gap sits almost entirely in the history reels, the ones with maps: 3.67 vs 4.50 passes and
2.67 vs 5.83 first-pass errors, against no difference on the worked problems or the English
explainers. But the curve is flat (Fig.~\ref{fig:curve}): within the lessons arm, passes and first-pass
findings show no trend over the 24 reels (Spearman $\rho = +0.03$, $-0.01$, $-0.02$) while the store
grew seventeenfold. Whatever a brief of lessons buys, it buys with the first handful; the fortieth
bullet adds nothing the agent acts on. Reading the 51 accumulated lessons says why: four of them are
narration-speed budgets that disagree with one another (3.7 and 3.9 characters per second, 2.5 and
3.0 words per second), several restate the same map-label rule from different reels, and none was
compiled into a check. Auto-acceptance grows a store; it does not grow a memory. That is the case for
the human gate and for L2 over L1: curation removes the duplicates and the contradictions, and only a
lesson turned into a rule is enforced on the fiftieth reel as surely as on the second.
\paragraph{And they do not move the judge.} Scoring all 48 reels with the absolute rubric gives the
two arms the same craft: judge 55.1 with lessons, 55.6 without (paired $p=0.53$; higher on 12 topics,
lower on 10), deterministic score 98.0 vs 97.7. Neither arm trends upward over the 24 reels; both
drift slightly down together ($\rho = -0.30$ and $-0.54$), which is the topic order, not the memory.
Twenty-four more topics say what the four of \S\ref{sec:ablation} said: the L1 tier changes how much
checking a reel needs, not what a judge of stills sees.

\subsection{Judge reliability}
\label{sec:reliability}
The round's verdict rests on one judge model voting once. To learn how much of it is the judge, we
re-ran the judging without making any reels: the absolute rubric three times on each of the sixteen
reels, the pairwise comparison three times per topic with the same judge, and once each with two other
judges (\texttt{claude-cli:haiku}, \texttt{claude-cli:opus}); about 170 judge calls in 45 minutes.

\paragraph{The absolute rubric is stable but sees little.} Within-reel standard deviation over three
re-scores has median 1.3 and maximum 2.7 on the rubric's 20--100 scale, so the flat score in
Table~\ref{tab:evolve} is not noise. Averaged over re-scores the rubric puts the candidate 3.5 points
ahead on the training topics (57.0 vs 60.5) and 1.9 behind on the held-out ones (59.5 vs 57.6), with
per-topic differences from $-9.9$ (the poem; baseline better) to $+8.7$ (the seasons). The rubric does
see the change; it does not agree with the pairwise verdict about its direction on holdout.

\paragraph{The pairwise verdict is repeatable---and biased by position.} Three repeats with the same
judge gave the candidate 8 of 8 topics every time, with 10--11 of 12 training votes and 10 of 12
held-out votes. But the breakdown by position (Table~\ref{tab:position}) shows that the judge picked
the candidate 48 of 48 times when it was shown \emph{second} and 14 of 24 times when it was shown
\emph{first}; over all votes it picked the second-shown reel 81\% of the time, and the two other judges
67\% and 62\%. Our vote design---three votes, order alternating from a fixed start---gave the candidate the second
slot twice per topic, which inflated its margin. Because the second-slot preference saturates (every
second-shown candidate won), the informative votes are the ones where the candidate was shown first:
8 of 12 on the training topics (two-sided binomial $p=0.39$) and 6 of 12 on the held-out ones
($p=1.0$), against 24 of 24 in the second slot ($p<10^{-6}$). The absolute rubric agrees: averaged
over re-scores and the eight topics the candidate is $+0.8$ points ahead, sign-flip permutation
$p=0.70$. There is no statistically supported evidence that the rewrite is better; the round's 11--1
was the slot. Balancing the vote count is not enough when the bias saturates---it still hands each side
two free votes per topic---so the harness now judges each side only on the votes where it was shown
first, its disadvantaged slot: the side with the higher first-slot win rate takes the topic, equal rates
tie. Under that rule the round is three topics won and one tied on train, two won and two tied on
holdout: it still passes its gate, with the uncertainty now visible in the ties rather than hidden in a
vote total.
\begin{table}[t]
\centering\small
\begin{tabular}{lccc|cc}
\toprule
 & \multicolumn{3}{c|}{candidate wins (sonnet, 3 repeats)} & \multicolumn{2}{c}{other judges, 1 run} \\
Topic & shown first & shown second & balanced & haiku & opus \\
\midrule
sci-rainbow & 0/3 & 6/6 & 0.50 & C & C \\
hist-chibi & 3/3 & 6/6 & 1.00 & C & C \\
solve-chicken-rabbit & 2/3 & 6/6 & 0.83 & C & C \\
en-seasons & 3/3 & 6/6 & 1.00 & C & C \\
sci-tides & 3/3 & 6/6 & 1.00 & C & C \\
hist-silk-road & 3/3 & 6/6 & 1.00 & B & C \\
solve-buoyancy & 0/3 & 6/6 & 0.50 & C & B \\
poem-jingyesi & 0/3 & 6/6 & 0.50 & B & B \\
\midrule
second-shown reel picked & \multicolumn{3}{c|}{58/72 = 81\%} & 67\% & 62\% \\
\bottomrule
\end{tabular}
\caption{Pairwise judging by position. ``Balanced'' is the mean of the two position-specific win
rates. C/B: topic winner by majority for the other judges. Topic-winner agreement between the five
judge runs is 0.82.}
\label{tab:position}
\end{table}

\paragraph{Where judges disagree, the human agrees with the minority.} The two other judges prefer the
baseline on the poem, and so does the absolute rubric by 9.9 points; one of us, before seeing any of
these numbers, had preferred the baseline's quieter staging too. Only the round's judge preferred the
candidate there---three to nothing, three times---and every one of those votes came from the second
slot. On the Silk Road and buoyancy topics one other judge each dissents. Agreement between judge runs
on the topic winner is 0.82; the disagreements concentrate on exactly the topics where the fact
coverage of Table~\ref{tab:facts} moved most.

\paragraph{A second round under the corrected gate.} With the position-robust rule in place we ran
the loop once more, from the round-1 skill. The proposer made two edits (35 lines: hero build recipes
for the frame worker; counting-figure and answer-box rules for worked problems) and declined a third.
On the training topics it won all four, judged from the disadvantaged slot (candidate-first votes 6 of
8; raw 14--2); on the held-out topics it won two and lost two (candidate-first 3 of 8; raw 8--8), and the
absolute composite moved $+3.4$ on train and $-2.7$ on holdout, with the tides reel falling from 60.0
to 48.7 on the rubric. The gate, which asks for wins $\geq$ losses on holdout, accepted it. The scripts
this time trade roughly five baseline facts (the 50-minute tide lag, Zhang Qian and 138\,BCE, the
buoyancy numbers) for roughly nine new ones (the 40$^\circ$ violet angle and the antisolar point,
spring tides, Buddhism on the Silk Road, Archimedes' principle); the round-1 pattern did not recur. The
second-shown reel was still picked 62\% of the time. We did not apply it: a 2--2 holdout and a rubric
drop on one topic are not a reason to change the playbook, which is the judgment the human gate exists
to make, and the paper reports the round as the gate's output rather than as an improvement.

Together with the flat absolute score, these support a conclusion reached elsewhere in other
settings \citep{notanoracle2026,moreconvincing2026,teachingmonster2026}: a learned judge is a useful
\emph{advisor} to a self-improving loop and a poor \emph{oracle}. The program view lets the loop keep
its oracles deterministic---rules, render checks and, in the benchmark, outcome tests on the
artifact---and reserve the judge for what only taste can rank.

\section{Related work}

\paragraph{Explainer and educational video from code.} TheoremExplainAgent \citep{theoremexplain2025}
and Code2Video \citep{code2video2026} generate Manim programs with planner, coder and critic agents
and judge them with LLM rubrics or a quiz; Paper2Video \citep{paper2video2025} adds pairwise
comparison against human talks. ManimAgent \citep{manimagent2026} is closest to us: a Manim-writing
agent with an episodic memory of success rationales and validated failure patterns grown from its own
task stream, evaluated by blind human Pass@1. Its memory is prose retrieved into the prompt; ours
compiles the checkable part into tested code, which is what makes retroactive audit and cross-model
portability possible. CineForge \citep{cineforge2026} and VideoWeaver \citep{videoweaver2026} evolve
policies or skills for long-form pixel video; what they learn cannot be run over the artifact,
because the artifact is not a program.

\paragraph{Experience memory for agents.} Reflexion \citep{reflexion2023}, ExpeL \citep{expel2024}
and ACE \citep{ace2026} store verbal reflections, insights or an evolving playbook; our L1 and L3
belong to this family, with ACE's incrementally curated playbook closest to \texttt{evolve}. Voyager
\citep{voyager2023} and ASI \citep{asi2025} store \emph{skills} as executable programs verified by the
environment; we store \emph{checks}, verified by their own examples, which is what a creative task
without an environment reward affords. The neuro-symbolic dual memory of \citet{neurosymbolicmem2026}
induces executable Python verifiers from failed trajectories and validates them on positive and
negative pools---the same contract as our L2, in household and web environments. SkillRevise
\citep{skillrevise2026} revises LLM-authored skills from execution traces.

\paragraph{Checkers from fixes.} KNighter \citep{knighter2025} synthesises static analysers from
historical kernel patches, validating each against the buggy and the patched code, and found 92
long-latent Linux bugs; test-driven checker development \citep{codechecker2024} makes the same
contract explicit. Our rules follow it exactly---must fail, must pass---but the corpus audited is the
agent's own past output, and the patches come from the agent's own revision log.

\paragraph{Self-modifying agents and judge reward hacking.} ADAS \citep{adas2024}, the Darwin G\"odel
Machine \citep{dgm2025} and GEPA \citep{gepa2025} let an agent rewrite its scaffold or prompts under
an objective score. \citet{notanoracle2026} and \citet{moreconvincing2026} show what happens when the
score is an LLM judge; \citet{fragility2026} show how sensitive such loops are to variance and task
order. Our evolve step is a small instance of this family with the guards those papers recommend: a
frozen holdout, pairwise rather than absolute judgment \citep{mtbench2023}, a deterministic layer the
proposer cannot edit, and a person at the end.

\section{Limitations}
\label{sec:limits}
\emph{Scale and family.} 111 reels, two evolution rounds, four topics per ablation cell, and one
model family (Claude) as director, judge and retrospective model; the three judges differ in size, not
provenance. \emph{No human preference study.} We cannot say whether any judge's taste matches viewers';
one author's eye agrees with the dissenting judges on one topic, which is an anecdote. \emph{Position
bias remains.} Judging from the disadvantaged slot discards the saturated votes but halves the sample;
under the balanced ablation design the second-shown reel was still picked 58\% of the time, so small
margins in pairwise verdicts carry little information. \emph{Null results at $n=4$.} The
ablation's ``lessons change nothing the judge sees'' and ``rules on are judged slightly worse'' are
within what position alone produces; they rule out large effects, not small ones. \emph{What rules can
see.} A rule checks program text. It cannot know that a topic requires naming an alliance or stating
the perihelion, that a label is too small on a phone, or that a mood is wrong; those remain lessons
(advice) or outcome tests (the companion benchmark). \emph{Rule provenance and precision.} Twelve of
thirteen rules were ported from the predecessor's checker, each from a real failure; one
lesson-to-rule path was exercised; one rule produced a false positive that the ablation exposed.
\emph{Human cost.} A person triaged 56 lessons and one evolve report and read sixteen scripts by
hand; we did not measure how that scales. \emph{Dependencies.} The free neural voices come from an
unofficial endpoint; commercial use needs a licensed provider, which the pipeline supports.

\section{Conclusion}
An explainer reel rendered from a program is a different object from a generated video, and the
difference is worth more than exact text: it makes the agent's own output something the agent, and
its authors, can check, audit and learn from at the level of code. \sys{} turns that into a
loop---log every pass, pair findings with fixes, propose lessons, compile the checkable ones into
tested rules, rewrite the playbook under a gate it cannot move---and a release-testing study shows
where each tier earns its place and where it does not. A rule found latent defects in reels made
before it existed; the same rules were almost never needed by a strong director and caught thirty mistakes a weak one
made;
lessons
bought fewer review passes without compounding and without moving a judge; and the preference
signal that accepted the one playbook rewrite turned out to be stable, repeatable, position-biased
and blind to the thirteen facts the rewrite traded for craft. Each of those findings came from
running a check over artifacts that already existed---reels, scripts, logs, votes---which is the
practice this paper argues for. The system, every rule with its examples, the lessons, the ablation
and reliability data and the evolve report are public at
\url{https://github.com/agentic-loops-x/Reels-RSI}.

\bibliographystyle{plainnat}
\bibliography{refs}

\appendix
\section{The thirteen rules}
\label{app:rules}
Each rule directory holds \texttt{rule.py}, at least one must-fail and one must-pass example;
\texttt{reels rules test} passes 13/13. Severity \emph{error} blocks render; \emph{warning} is
reported.

\begin{table}[h]
\centering\small
\begin{tabular}{llp{9.6cm}}
\toprule
Rule & Severity & What it catches \\
\midrule
\texttt{timeline-registration} & error & a sub-composition that does not register its paused timeline under its composition id \\
\texttt{nondeterminism} & error & \texttt{Math.random}, \texttt{Date.now}, \texttt{requestAnimationFrame}, timers or self-driving animation in a seek-rendered frame \\
\texttt{cjk-font} & error & CJK text without an \texttt{@font-face} for a shipped font (the render browser has no CJK system fonts) \\
\texttt{dasharray-attr} & error & a \texttt{stroke-dasharray} attribute in the markup of a path that a tween draws on \\
\texttt{svgorigin-one-sided} & error & \texttt{svgOrigin} given only in a \texttt{fromTo}'s to-vars \\
\texttt{negative-zindex} & error & a negative z-index, which puts the layer behind the composition root \\
\texttt{caption-band} & warning & frame text placed in the band reserved for karaoke captions \\
\texttt{chained-no-position} & warning & a chained \texttt{.to()} without a position argument, which drifts to the timeline's end \\
\texttt{repeat-fromto} & warning & a second \texttt{fromTo} on the same target without \texttt{immediateRender:false} \\
\texttt{round-cap-mask} & warning & a reveal mask with round caps, which leaks a dot before the stroke begins \\
\texttt{polygon-winding} & warning & a counter-clockwise ring in d3-geo map data (``the globe minus this shape'') \\
\texttt{unpinned-cdn} & warning & a CDN script without a version pin \\
\texttt{visible-from-state} & warning & a \texttt{fromTo} that starts visible and fades out without \texttt{immediateRender:false} (learned in this study) \\
\bottomrule
\end{tabular}
\caption{Rules shipped with \sys{} 0.1. A user's own rules live beside them and may override a
built-in of the same name.}
\end{table}

\section{Accepted lessons}
\label{app:lessons}
The fifteen lessons a person accepted during the study, as injected into frame briefs (abridged).
\begin{itemize}
\item \textsc{frame} --- A tween starting from a visible state on an element meant to appear later needs \texttt{immediateRender:false} (\emph{rule}).
\item \textsc{frame} --- Labels inside coloured regions go at the region's widest part and get a board-coloured outline.
\item \textsc{frame} --- Underlines, boxes and circles that mark a piece of text ride on the text element, never on guessed coordinates.
\item \textsc{frame} --- Draw chalk boxes from four straight wobbly segments; curve-smoothing turns a rectangle into a blob (\emph{when preset=chalk}).
\item \textsc{frame} --- On 9:16 canvases scale the hero to the full width and use the lower half for the working (\emph{when aspect=9:16}).
\item \textsc{frame} --- Size on-figure labels at 28~px or larger; keep construction lines heavy and inside the figure.
\item \textsc{frame} --- Pick label colours as pre-lightened variants that pass contrast on the board.
\item \textsc{frame} --- Highlight a named composite shape by filling its parts, not by tracing a thin outline; put the answer box beside it.
\item \textsc{frame} --- A hero word goes in an empty region of the hero, never over bright strokes.
\item \textsc{frame} --- Clamp computed rays so endpoints and labels land inside the canvas above the caption band.
\item \textsc{director} --- Budget about 3.9 Chinese characters per second for the female teaching voice, plus 10\% when reading letter names.
\item \textsc{director} --- Budget about 4.5 characters per second, pauses included, for the male narrator.
\item \textsc{director} --- Spread each frame's build-up across its narration; single out the key shape on the word that names it.
\item \textsc{solve} --- Give every reel a drawing kit with the figure computed from coordinates (\emph{kit}).
\item \textsc{solve} --- For a problem from a photo, transcribe text and figure, restate them for confirmation, and build the figure from the given lengths, never from pixel proportions.
\end{itemize}

\end{document}